\def\CLASSINPUTinnersidemargin{0.75in}
\def\CLASSINPUToutersidemargin{0.75in}
\documentclass[letterpaper, 10pt, conference]{IEEEtran}
\IEEEoverridecommandlockouts
\usepackage[hidelinks,breaklinks]{hyperref}
\usepackage{cite}
\usepackage{amsmath,amssymb,amsfonts}
\usepackage{algorithm}
\usepackage{algpseudocode}
\usepackage{graphicx}
\usepackage{textcomp}
\usepackage{xcolor}
\def\BibTeX{{\rm B\kern-.05em{\sc i\kern-.025em b}\kern-.08em
    T\kern-.1667em\lower.7ex\hbox{E}\kern-.125emX}}
\renewcommand{\IEEEtitletopspaceextra}{18pt}
\begin{document}
\title{Socially-Aware Autonomous Doorway Traversal and Payload Delivery for Emergency Assistance}

\author{
    Andrew Snowdy, Ananya Trivedi, Sarvesh Prajapati, Lorena Maria Genua, and Ta\c{s}k{\i}n Pad{\i}r$^1$%
    \thanks{The authors are with the Department of Electrical and Computer Engineering, Northeastern University, Boston, MA 02115, USA. Emails: \{\url{snowdy.a}, \url{trivedi.ana}, \url{prajapati.s}, \url{genua.l}, \url{t.padir}\}@northeastern.edu}%
    \thanks{$^1$Ta\c{s}k{\i}n Pad{\i}r holds concurrent appointments as a Professor of Electrical and Computer Engineering at Northeastern University and as an Amazon Scholar. This paper describes work performed at Northeastern University and is not associated with Amazon.}%
    \thanks{This research was funded, in part, by the Advanced Research Projects Agency for Health (ARPA-H) Agreement No. 140D042590012. The views and conclusions contained in this document are those of the authors and should not be interpreted as representing the official policies, either expressed or implied, of the U.S. Government.}%
}

\maketitle

\begin{abstract}
In this work, we focus on the scenario of a robot-assisted emergency evacuation. We consider two capabilities relevant to such a setting. The first is opening doors ahead of the people being evacuated, so that their path toward an exit stays clear. The second is retrieving rescue equipment and delivering it to the emergency responders carrying out the evacuation. From a systems perspective, this involves several tasks at once. The robot must locate ADA-compliant door buttons and the rescue equipment it needs to retrieve. Additionally, it must remain aware of the people around it and adapt its behavior to them, so that it supports the evacuation rather than getting in the way. We address these demands with a behavior tree at the core of our framework. This structure is chosen for its ability to select high-level tasks based on environmental triggers, and to extend to new situations as they arise. We evaluate the system in 105 trials on the Toyota Human Support Robot, across four hardware and three simulation scenarios. These trials capture the decisions the robot must make in this setting: whether to press a door button, yield to a nearby person, walk through a door someone else is holding, or first retrieve rescue equipment before traversing the door. Overall, the system completes 97 of the 105 trials successfully. These results suggest our framework provides a practical basis for robotic assistance in broader emergency response tasks. Code and video demonstrations are available at \url{https://github.com/AndrewSnowdy/hsr_mm_control}.
\end{abstract}

\section{Introduction}
\label{sec:introduction}
The July 2025 Gabriel House fire in Fall River, Massachusetts~\cite{gabrielfire} claimed the lives of ten residents of an assisted living facility. Many of the victims had limited mobility and were unable to move through narrow hallways once the primary elevator failed. In high-stress events like this, infrastructure such as ADA-compliant doors can turn into bottlenecks. They are slow to open, may require a button press, and can force a queue to form as people wait to pass through one at a time. Evacuating people safely under these conditions is difficult, highlighting a clear need for systems that can assist an evacuation as it unfolds. 

\begin{figure}[t]
    \centering
    \includegraphics[width=0.8\linewidth]{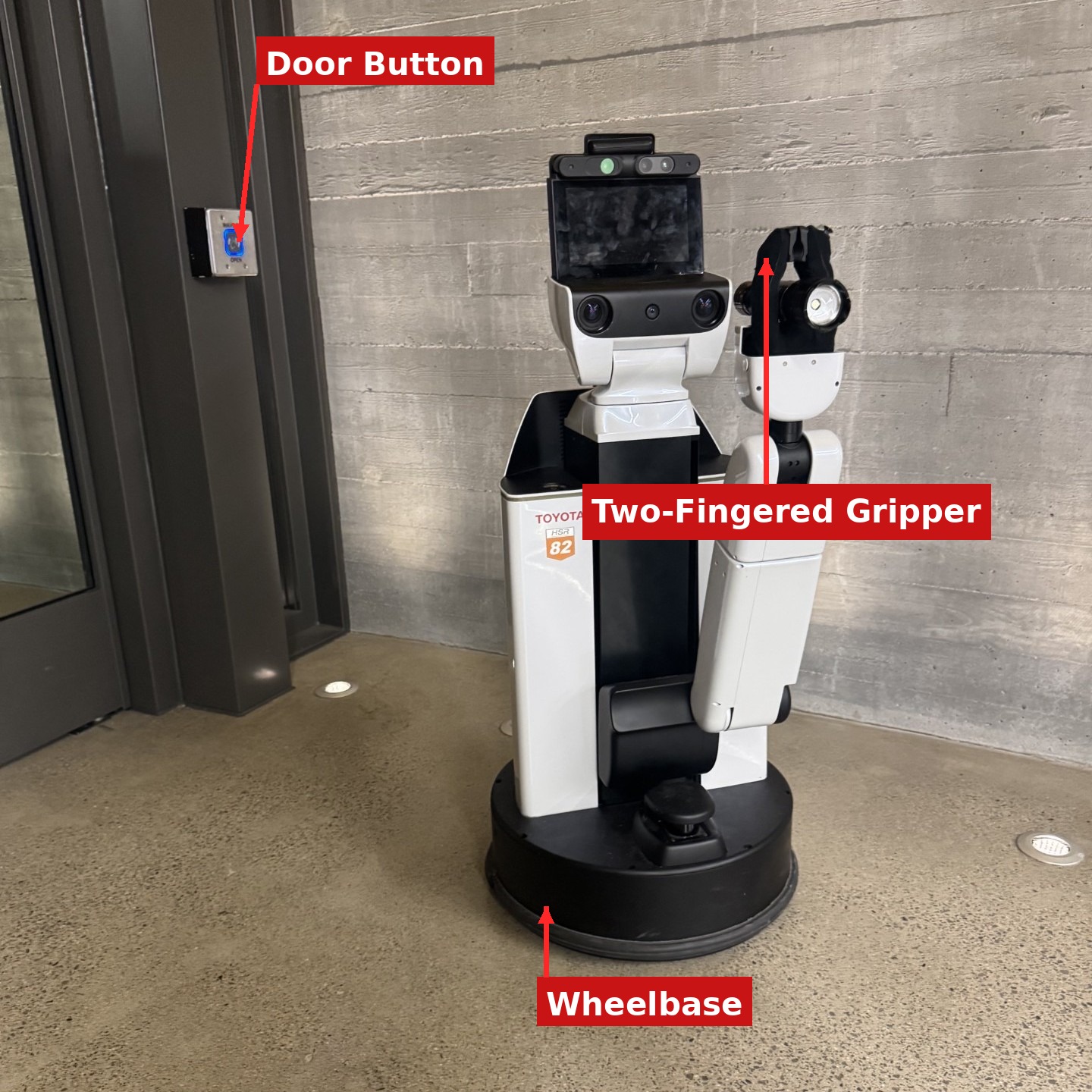}
    \caption{Toyota Human Support Robot (HSR) used in our experiments. The two-fingered gripper grasps payloads such as water bottles and flashlights and presses the ADA-compliant door button, while the omnidirectional wheelbase provides planar mobility.}
    \label{fig:robot_near_door}
\end{figure}

In this work, we investigate how indoor service robots can help with tasks like these. Our focus is on clearing pathways for people as a rescue unfolds. If a robot can keep doorways open while people follow behind, it can save valuable time. This requires the robot to interact with surrounding people by leading the way when the path is clear, yielding to those showing intent to pass, and proceeding when someone holds the door open. Additionally, the robot can assist the emergency personnel on site, who are occupied with the evacuation itself. When the environment is known, it can fetch requested items and bring them to these responders. Together, this establishes a framework where the robot both helps lead people out and supports the personnel carrying out the rescue.

From a systems standpoint, this calls for a modular design. The perception module maintains a map of the surroundings and localizes the robot within it. It must also detect the objects that matter for the task, including ADA-compliant door buttons and the rescue equipment to retrieve. A separate module infers the intent of the people nearby, so that the robot can interact with them. From these inputs, the robot selects a high-level task, which is then passed to a motion planning module. This module translates the task into base and arm motion, so that the robot can navigate through the space and pick up objects. Finally, as the deployment setting or the robot changes, we may need to alter existing behaviors or add new ones.
\begin{figure*}[t]
    \centering
    \includegraphics[width=\linewidth]{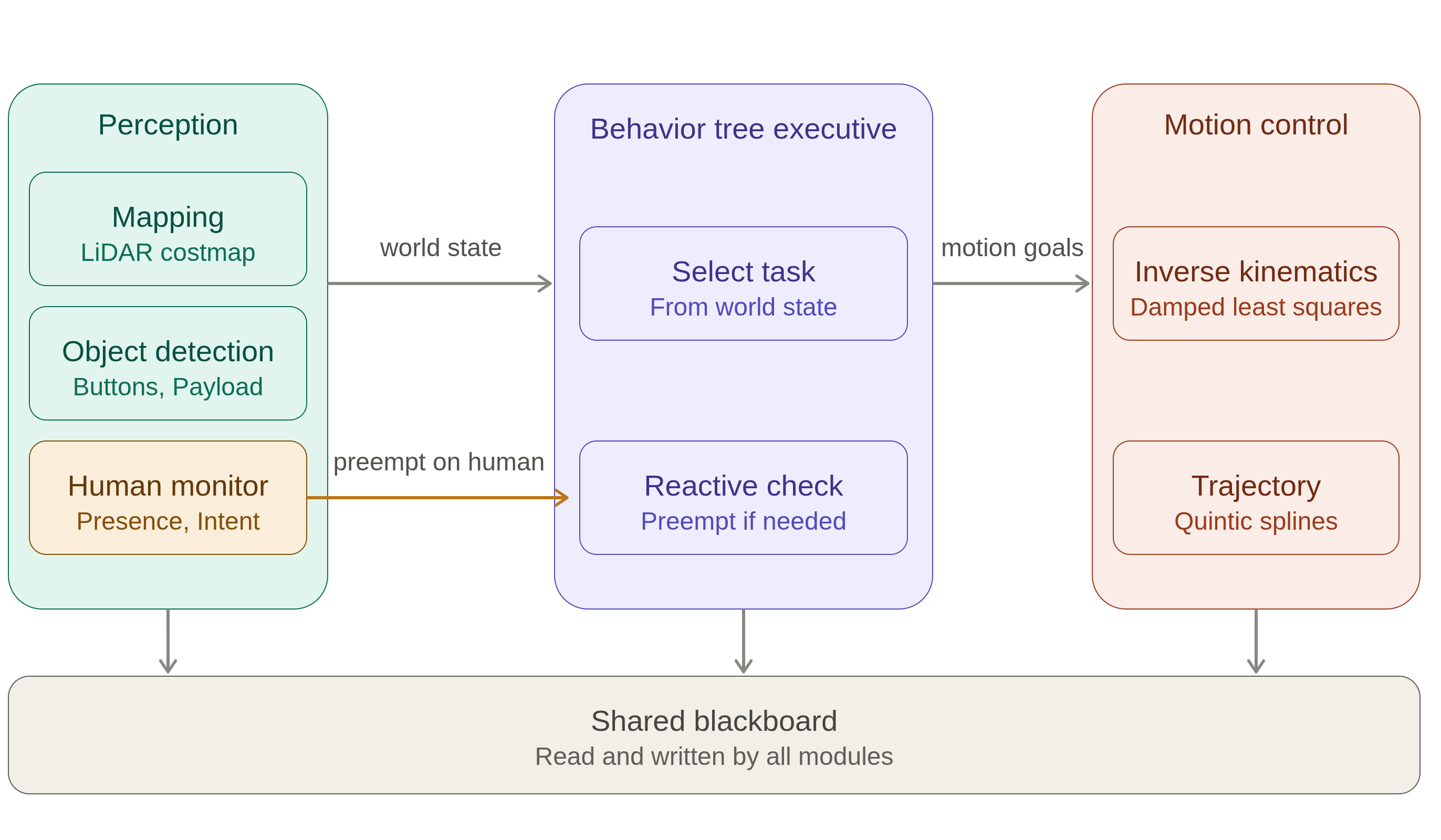}
    \caption{Overview of the proposed modular architecture. The Perception module builds a LiDAR costmap and localizes the robot. It also detects task-relevant objects such as door buttons and the payload, along with nearby people and their intent. These outputs form the world state. The Behavior Tree executive uses this state to select the current task and issue motion goals. The Motion control module realizes these goals through a damped least squares inverse kinematics solver and quintic spline trajectories. All modules communicate through a shared Blackboard. The orange arrow marks the reactive interrupt path. When a person enters the workspace, the Reactive check node preempts the active task so the robot yields.}
    \label{fig:pipeline}
\end{figure*}
Behavior trees~\cite{colledanchise2018behavior} are well suited to a problem with this structure. A behavior tree organizes a task as a hierarchy of nodes. The internal nodes select which action to run based on the current state of the environment, while the leaf nodes issue the actual motion commands. In our setting, for instance, an internal node can track whether a person is moving to pass through the door. When that happens, it switches the robot to a higher-priority action, so the robot yields instead of continuing through. The leaf node below then issues the velocity command that carries out the action. This structure also lets us add new behaviors as a new subtree. We can do this without changing the nodes already in place, which gives us the extensibility we need as settings or the robot change~\cite{faconti2021behaviortree}.

In this paper, we cast robotic evacuation assistance as a concrete set of decisions the robot must make at a doorway, spanning both clearing pathways for the people being evacuated and delivering equipment to responders. We design a modular behavior tree framework that ties perception, human intent inference, and task selection into a single reactive system, structured so that behaviors can be extended as the deployment setting or robot changes. We validate the framework on the Toyota Human Support Robot, shown in Figure~\ref{fig:robot_near_door}, in both hardware and simulation. We characterize where the system succeeds and identify where perception remains a limiting factor for future improvement. Figure~\ref{fig:pipeline} shows an overview of the proposed pipeline.

\section{Related Work}
Several prior systems have tackled parts of the autonomous door traversal problem. Chitta et al.~\cite{chitta} developed a motion planning system for coordinating a mobile robot's base and arm to push and pull doors open. Arduengo et al.~\cite{arduengo} addressed the challenge of operating different types of doors by learning the physics of door opening in real time and adapting the robot's handle grasping strategy accordingly. Both of these methods, however, did not consider navigation to and through the door. From a social navigation perspective, Thomas et al.~\cite{messing2018after} studied how robots and humans can negotiate right-of-way at doorways, though only for passing through already-open gaps without any door manipulation.

\begin{figure}[!t]
    \centering
    \includegraphics[width=1.\linewidth]{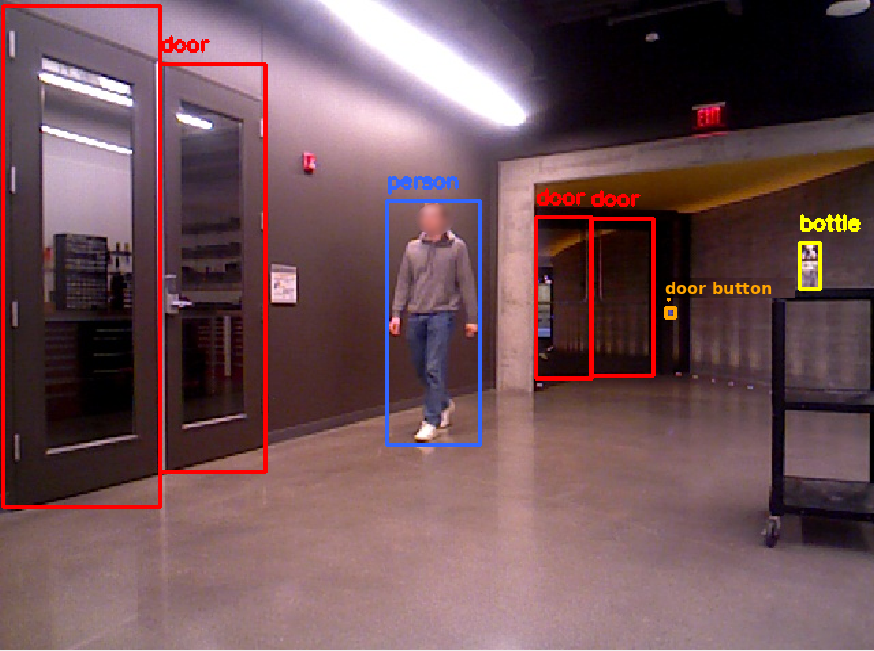}
    \caption{Real-time perception using YOLOv11n. The system identifies ADA-compliant buttons, people, bottles and doors to inform the Behavior Tree's social navigation logic.}
    \label{fig:yolo_example}
\end{figure}
On legged robots such as quadrupeds~\cite{quadruped}, Sleiman et al.~\cite{sleiman2023versatile} and Zhang et al.~\cite{marco_door_opening} demonstrated autonomous door traversal by having the system determine when the robot should make or break contact with the door and where on the door surface to do so. However, all planning is performed ahead of time, and if the robot is disturbed during execution a full replan is required online, adding significant time before the robot can continue its  motion.  More recently, Calvert et al.~\cite{ihmc} demonstrated door traversals on a bipedal humanoid using a behavior tree that executes walking and arm movements simultaneously.

While each of these systems advances a specific aspect of the problem, none integrates door manipulation with the ability to autonomously detect and react to humans sharing the doorway space. Our work addresses this gap by combining object detection, door manipulation, and human-reactive decision making into a single modular system architecture whose individual components can be independently extended or replaced. The modular system architecture allows the robot to quickly replan and adapt when unexpected changes in the environment prevent the current plan from being executed.
\section{System Architecture}
Following the modular structure motivated in Section~\ref{sec:introduction}, we organize the system into three layers: a perception stack that estimates the world state, a behavior tree executive that selects the active task from that state, and a motion control layer that executes it.
\subsection{Perception Module}
The perception layer supplies the world state that the behavior tree reasons over. The 2D LiDAR builds a costmap and localizes the robot within it. The head-mounted RGB-D camera handles object detection, for which we use YOLOv11n~\cite{yolov11n}, a lightweight real-time detector that returns labeled bounding boxes.

The pretrained YOLO detector is reliable on common classes like people and bottles, but not on the ADA-compliant door buttons and glass doors the task depends on. We therefore fine-tuned a second instance for these, using 600 training and 150 validation images that we collected and annotated to span the button styles and door configurations in our environment. The two YOLO models run in parallel. This way the prediction accuracy of every class the behavior tree relies on is maintained, each served by a model trained specifically for it. Figure~\ref{fig:yolo_example} shows an example of the detector running in our environment, with the door buttons, doors, people, and bottles labeled.

Due to the HSR's limited onboard compute, running inference on the platform itself at the rates needed for real-time decision-making is not feasible. Inference is therefore offloaded to an external NVIDIA Jetson Orin NX. This delivers detections at approximately 8~Hz, each available within 200~ms of the frame being captured. In our experiments we found this to be well within the timescales of human-interactive autonomous doorway navigation.

The raw detections are refined on the HSR's onboard computer before the behavior tree uses them. Each is filtered by a confidence threshold. Surviving detections are then tracked across frames by matching those at nearby positions. This suppresses transient false positives and confirms that a detection persists rather than flickering in and out. 

The refined detections must then be placed in the map. For most classes this uses depth: the center of the bounding box is projected through the RGB-D depth image to recover a 3D position. For glass doors the depth returns are unreliable, so here we combine the camera and LiDAR sensors. The YOLO bounding box picks out the door's angular slice in the LiDAR scan. Fitting a line to the returns in that slice then recovers the door's plane and orientation from the LiDAR rather than from the depth image.

\subsection{Behavior Tree Executive}
The core decision-making logic is implemented using the \textit{BehaviorTree.CPP} library~\cite{faconti2021behaviortree}. A classical finite state machine can only move between states along transitions defined ahead of time, so reacting to a new event requires being in a state that already anticipates it. A behavior tree instead re-evaluates its priorities on every tick and can drop a lower-priority action for a higher-priority one whenever conditions change. 

\begin{figure}[t]
    \centering
    \includegraphics[width=\linewidth]{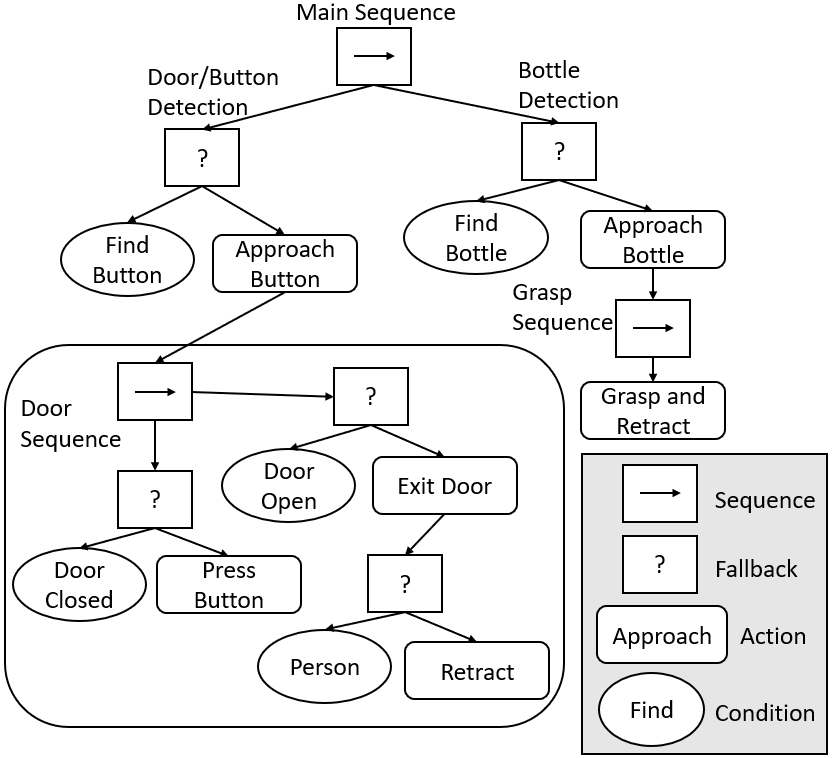}
    \caption{The behavior tree used by the executive. Selector and Sequence nodes at the upper levels arbitrate between tasks based on the world state, while the leaf nodes below issue the button-press, traversal, grasp, and retract commands. The legend at the lower right defines each node type.}
    \label{fig:behavior2}
\end{figure}

This is the reactivity our setting needs. A high-level Selector node continually monitors for people, and if one enters the defined workspace the tree immediately preempts the active manipulation task and drops to a safe standby state.
Several other contextual behaviors are embedded in the tree to make use of the perception inputs. One evaluates the button type, distinguishing a push button from a proximity plate, to set how far the end-effector must travel. Another assesses whether the door is already open, which determines whether a button press is needed at all. A third estimates the velocity of a nearby person, classifying them as either an obstruction to yield to or an agent holding the door open. Figure~\ref{fig:behavior2} shows the full behavior tree, with the Selector nodes at the top that arbitrate between tasks and the leaf nodes below that issue the actual button-press, traversal, and retract commands. Section~\ref{sec:experimental_results} details how each of these behaviors is structured, the rules they execute, and the conditions that trigger them.

\subsection{Mobile Manipulation Module}

Once the Behavior Tree executive triggers a high-level task, the motion control layer carries it out on the HSR's omnidirectional base and 5-DOF arm. It runs in two modes. During navigation the arm is held tucked against the body, which keeps the robot compact and stable while the base drives to a target. Once there, the manipulation mode takes over, and a Damped Least Squares (DLS) inverse kinematics (IK) solver computes the arm joint angles that bring the end-effector to the target pose, such as a door button or the payload. Smooth motion in both modes is produced by a quintic spline trajectory controller.
\subsubsection{Inverse Kinematics via Damped Least Squares}
Following~\cite{buss2004introduction}, the joint update at each iteration is:
\begin{equation}
    \dot{\mathbf{q}} = \mathbf{J}_t^\top \left( \mathbf{J}_t \mathbf{J}_t^\top 
    + \lambda^2 \mathbf{I} \right)^{-1} \mathbf{e}
\end{equation}
\noindent where $\dot{\mathbf{q}}$ is the joint velocity update at this iteration, $\mathbf{J}_t$ is the Jacobian mapping joint motion to end-effector motion, $\mathbf{e}$ is the error between the current and target end-effector pose, $\mathbf{I}$ is the identity matrix, and the damping factor $\lambda = 0.5$ keeps the step stable near singular configurations. Each update is scaled by a step size $\alpha = 0.1$, with every joint clamped to $|\Delta q_i| \leq 0.1$~rad per iteration to keep motion smooth. The solve stops once the end-effector error falls below $\epsilon = 10^{-3}$~m, or after at most 200 iterations.

The solver is warm-started from the previously converged configuration, so each solve begins close to the answer and reduces to a small local refinement, converging in approximately 20~ms. An optional wrist-leveling constraint enforces $q_{\text{wrist}} = -\pi/2 - q_{\text{arm\_flex}}$, where $q_{\text{arm\_flex}}$ is the arm's pitch joint angle and $q_{\text{wrist}}$ the wrist joint angle. This ties the wrist to the arm's pitch so the wrist counter-rotates as the arm raises or lowers, keeping the gripper level. It is used for flat-palm tasks, such as carrying a retrieved payload.

\begin{table}[t]
\centering
\caption{System Latency and Timing Characteristics}
\label{tab:timing}
\begin{tabular}{|l|c|}
\hline
\textbf{Subsystem Component / Metric} & \textbf{Measured Temporal Value} \\
\hline
YOLO Inference Latency                        & $\sim$125~ms \\
Perception End-to-End Tracking Latency          & $<$200~ms \\
IK Solver Execution Time (Warm-started)         & $\sim$20~ms \\
Behavior Tree Update Interval                   & 100~ms \\
\hline
Total Reactive Preemption Latency              & $<$200~ms \\
ADA Automatic Door Timeout Window               & $\sim$10.0~s \\
\hline
\end{tabular}
\begin{flushleft}

\end{flushleft}
\end{table}
\subsubsection{Quintic Spline Trajectory Generation}
To keep motion smooth and predictable around people, both base and arm movements follow time-parameterized quintic polynomials~\cite{quintic_splines}. These splines enforce zero velocity and zero acceleration at both endpoints of a motion, so the robot eases in and out of every move rather than starting or stopping abruptly, avoiding sudden changes in momentum that could be hazardous to anyone nearby.

Since the base translates while the arm articulates, the two are synchronized to reach the target at the same time. The duration of a motion is set online from the base travel distance and a minimum-duration floor:
\begin{equation}
    T = \max\!\left(\frac{L}{v_{\text{target}}},\; T_{\text{min}} \right)
\end{equation}
\noindent where $L$ is the base translation distance and $v_{\text{target}}$ the target linear velocity. Under nominal motion $v_{\text{target}} = 0.2$\,m/s. While passing through the door, however, the robot must clear it before the ADA auto-close window of roughly 10\,s expires, so the target velocity is raised to $v_{\text{target}} = 0.4$\,m/s for that segment. The floor $T_{\text{min}} = 1.5$\,s keeps very short motions, such as small terminal adjustments during a button press, from being scaled over too brief a horizon and producing high accelerations near people.

A feedforward-PD controller ($K_p = 3.0$, $K_d = 0.01$) tracks the resulting spline at 100\,Hz. A software watchdog runs alongside it and commands zero velocity to all joints if the position error exceeds 0.5\,m or the yaw error exceeds 0.5\,rad. The timing characteristics of these subsystems, together with those of the perception pipeline, are summarized in Table~\ref{tab:timing}.

\section{Experimental Results}
\label{sec:experimental_results}
 
To evaluate the proposed Behavior Tree (BT) architecture, we conducted a total of 105 experimental trials across seven scenarios: four hardware trials conducted on the physical HSR platform and three simulation trials executed in the Gazebo simulator. Each scenario was repeated multiple times to yield statistically meaningful success rates. The four hardware scenarios are denoted S1–S4, where S4 is split into S4a (water bottle retrieval) and S4b (flashlight retrieval). The simulation scenarios are denoted S5–S7. The success rates are summarized in Table~\ref{tab:results}.
 
\begin{table}[t]
\centering
\caption{Experimental Trial Success Rates}
\label{tab:results}
\begin{tabular}{|l|c|c|c|}
\hline
\textbf{Scenario} & \textbf{Type} & \textbf{Trials} & \textbf{Success Rate} \\
\hline
S1: Nominal Traversal       & HW  & 15 & 15/15 (100\%) \\
 S2: Human Obstruction       & HW  & 15 & 14/15 (93\%) \\
S3: Assisted Traversal      & HW  & 15 & 13/15  (87\%)  \\
S4a \text{\&} 4b: Pre-Traversal Manipulation   & HW  & 30 & 26/30  (87\%)  \\
\hline
S5: Full Autonomy Chain     & SW  & 10 & 10/10  (100\%)  \\
S6: Human-Induced Preemption & SW & 10 & 10/10 (100\%) \\
S7: Static Obstacle + Assisted & SW & 10 & 9/10 (90\%) \\
\hline
\end{tabular}
\end{table}

\subsection{Hardware Scenarios}
\begin{figure*}[t]
    \centering
    \includegraphics[width=\linewidth]{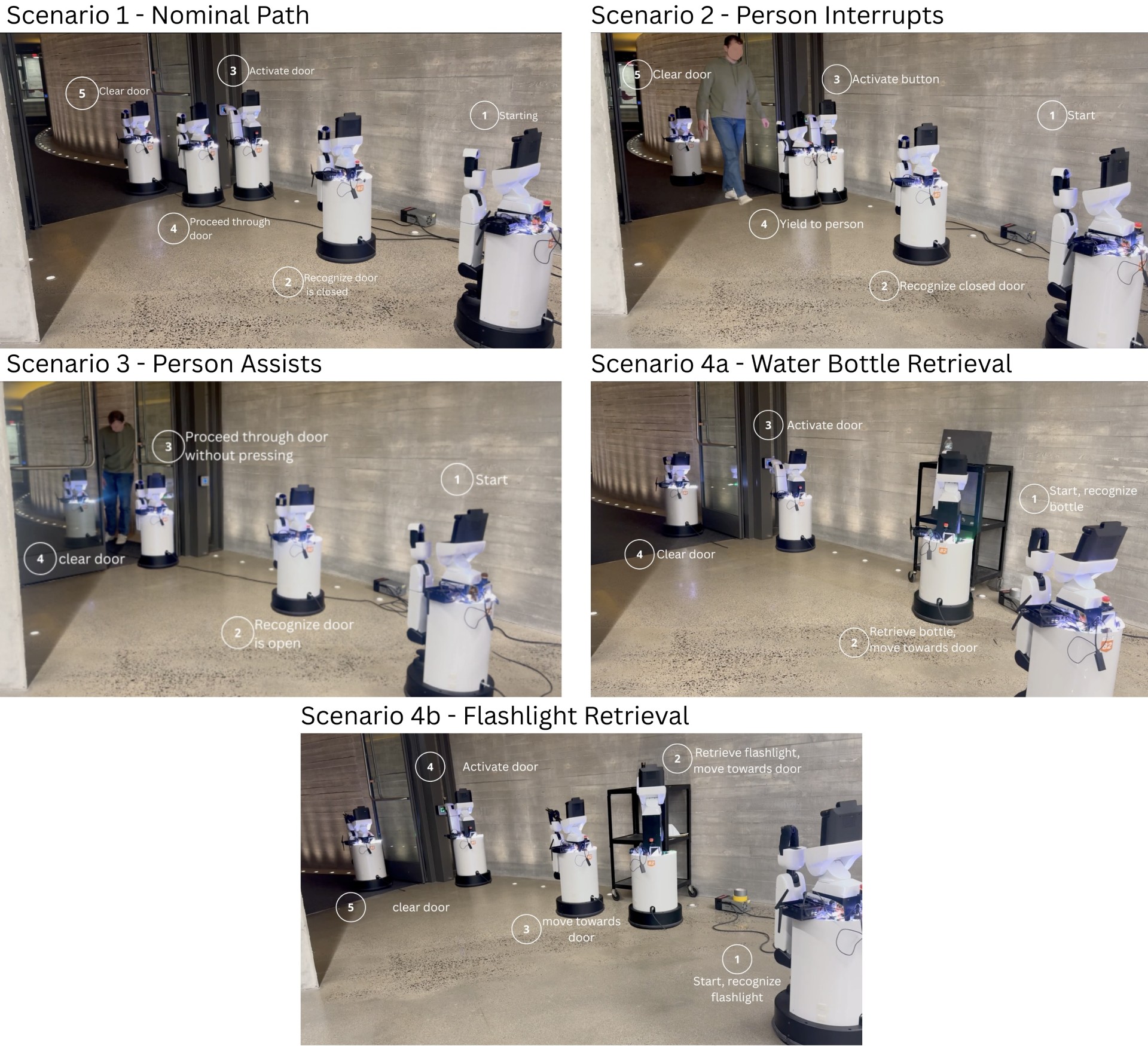}
    \caption{Sequential frames from the four hardware scenarios: nominal traversal (S1), yielding to a person crossing the doorway (S2), passing through a human-held door (S3), and payload retrieval before traversal with a water bottle (S4a) and a flashlight (S4b).}
    \label{fig:hardware}
\end{figure*}

 Figure~\ref{fig:hardware} shows sample frame sequences from each hardware scenario, tracing the robot from its start pose through button press, yielding, or payload grasp to final door traversal. We now discuss each scenario in further details. 
\subsubsection{S1: Nominal Autonomous Traversal (15/15)}
In the baseline scenario, the HSR operated without any human agents present. 
The robot successfully localized the ADA-compliant proximity button in all fifteen trials using 
the fused YOLOv11n and LiDAR pipeline, executed the quintic spline reach-and-press 
sequence, and navigated through the doorway within the ADA auto-close window of 
approximately 10 seconds. This result confirms that the core perception-to-manipulation 
pipeline is reliable under nominal conditions, with the IK solver consistently converging 
in under 200 iterations and completing in approximately 20 ms after warm-starting from 
the robot's current joint configuration.

\subsubsection{S2: Dynamic Human Obstruction (14/15)}
This scenario tested the BT's social preemption logic. When a human agent walked 
through the doorway during the robot's active approach, the system successfully halted 
its manipulation sequence and retracted to a safe standby position in all but one trial. 
Human detection was handled by the general-purpose YOLO inference stream; upon 
classification as a \textit{person} within the defined ground-plane proximity zone, 
the BT's \texttt{IsPathClear} condition node returned failure, immediately preempting 
the active \texttt{ExitDoor} subtree. Person state was maintained across frames using 
an exponential smoothing filter ($\alpha = 0.3$) on both position and velocity, with 
trajectory prediction extrapolated forward across four steps at 0.2 s intervals
providing the BT with a short-horizon occupancy forecast of the doorway region rather 
than a purely reactive point estimate. The single failure occurred when a fast-moving 
human cleared the proximity zone before the retract sequence completed, causing the 
robot to briefly resume approach before re-detecting the obstruction.
 
\subsubsection{S3: Assisted Traversal (13/15)}
The assisted scenario—wherein a human holds the door open and the robot is expected 
to bypass the button-press sequence and proceed directly through the open doorway
proved to be the most perceptually demanding hardware scenario, achieving a 87\% 
success rate. Failures in the two trials were attributable to a consistent edge case in the 
door perception pipeline: the door was initially detected as closed at the start of each 
trial (which is the correct assumed state), but the transition to an ``open'' classification 
required the door to move sufficiently out of the camera's central field of view so that 
the LiDAR-based normal estimation could no longer detect a planar obstacle. In trials 
where the human held the door only partially open, the door remained partially in-frame 
and the system continued to classify it as closed, triggering the button-press sequence 
erroneously. This reveals a fundamental limitation of the current perception architecture 
when handling intermediate door states.

\subsubsection{S4a \& S4b: Pre-Traversal Mobile Manipulation (26/30)}
Prior to a door traversal, the robot was required to identify and grasp a 
target object (a water bottle or flashlight) in the same room. This scenario achieved an 87\% success 
rate. The four failures were attributable to object detection rather than manipulation 
execution: in these cases, the YOLOv11n model failed to provide a stable, high-confidence 
detection of the target object, either due to partial occlusion by environmental clutter or 
suboptimal lighting conditions relative to the training distribution. When detection was 
stable, the DLS IK solver and quintic spline trajectory generator produced reliable and 
smooth grasps, indicating that the manipulation stack itself is robust given accurate 
perception input.
 
\subsection{Simulation Scenarios}

\begin{figure*}[t]
    \centering
    \includegraphics[width=\linewidth]{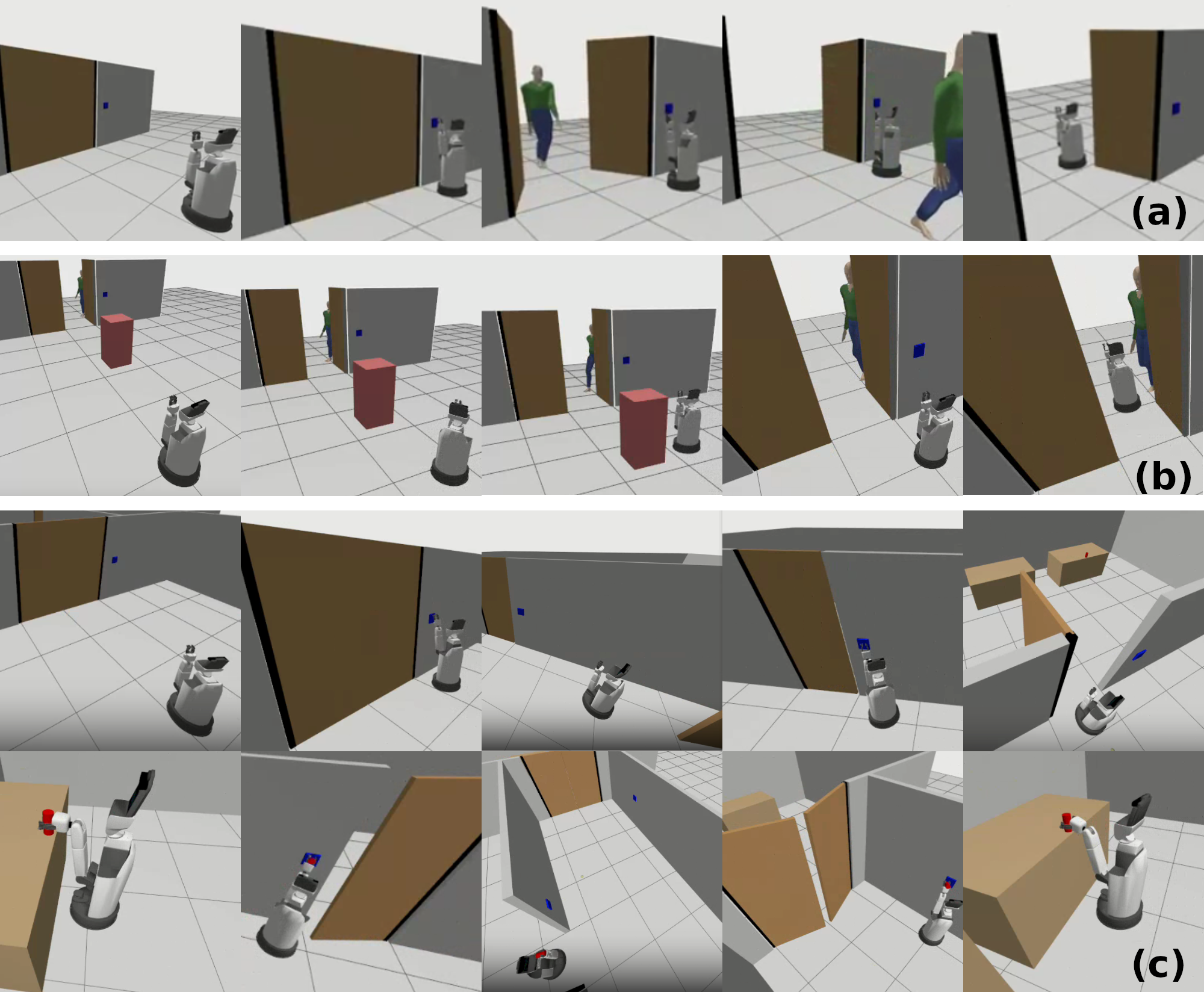}
    \caption{Sequential frames from the three simulation scenarios: yielding to a simulated person crossing the doorway (S6), navigating around a static obstacle before passing through a human-held door (S7), and the full autonomy chain of traversing two doors, retrieving a bottle, and returning to deposit it (S5).}
    \label{fig:simulation}
\end{figure*}

Figure~\ref{fig:simulation} shows sample frame sequences from each simulation scenario, following the robot through preemption, obstacle avoidance, and the full retrieve-and-return chain across the office environment. 
\subsubsection{S5: Full Autonomy Chain (10/10)}
To validate the system's capability for extended, task-level operation, a simulation 
scenario was constructed in which the HSR autonomously navigated through an office 
environment containing two sets of doors in differing configurations, retrieved a target 
bottle from a room at the far end, and returned through both doorways to deposit the 
bottle on a table in the starting room.  This scenario demonstrates that the BT architecture 
composes correctly across multiple sequential door-traversal instances and retains 
manipulation precision following complex navigation sequences.
 
\subsubsection{S6: Human-Induced Preemption in Simulation (10/10)}
Mirroring the hardware obstruction scenario, a simulated human agent was introduced 
to intersect the robot's planned path through the doorway. The BT executed the retraction 
sequence correctly in all ten trials. The ground-plane bounding box heuristic—wherein 
a person classified within a predefined spatial region relative to the door frame triggers 
a retreat to the last safe waypoint—proved sufficient for reliable preemption without 
requiring explicit velocity-based trajectory prediction in the simulation environment.
 
\subsubsection{S7: Static Obstacle Avoidance with Assisted Entry (9/10)}
This scenario combined two behavioral demands: avoidance of a static obstacle placed 
along the nominal approach path, and recognition of a human-held open door that 
eliminates the need for button interaction. The bug algorithm~\cite{bug_algorithms} successfully guided the 
robot around the obstacle in all ten trials, while the costmap-integrated door state 
estimator correctly classified the door as open in nine of ten cases, allowing the BT to 
skip the button-press subtree. One failure occurred when the obstacle's placement 
created a narrow corridor that caused the costmap inflation layer to temporarily block 
the door threshold, preventing the robot from planning a valid traversal path.
 
\section{Discussion}

The experimental results show that the proposed Behavior Tree (BT) architecture provides a flexible framework for socially-aware door navigation. The nominal traversal scenario reached 100\% success in hardware, while the human-obstruction scenario achieved 93\% (14/15) in hardware and 100\% in simulation. This confirms that the core BT logic and the IK stack operate reliably when perception inputs are stable and unambiguous. The system preempts active manipulation in under 200 ms and retreats to a defined safe state. In a BT, this preemption follows from the tree structure itself, since a higher-priority node can take over as soon as its condition holds. A Finite State Machine (FSM) would require an explicit transition from each state where preemption is allowed, which becomes harder to keep consistent as the behavior library grows.
 
\subsection{Perception as the Primary Failure Mode}
The most salient finding across all scenarios is that perception robustness, rather than 
motion planning or manipulation, is the dominant source of system failures. Both the 
assisted traversal scenario and the post-traversal manipulation scenario 
failures were exclusively due to detection errors rather than downstream kinematic or planning 
issues. This pattern highlights a structural dependency of the BT executive on the quality 
of its perceptual inputs: when detections are stable and confident, the downstream 
manipulation pipeline performs reliably.
 
The door state estimation problem is particularly instructive. The current approach 
assumes a binary open/closed classification derived from the presence or absence of a 
planar obstacle in the LiDAR scan. This assumption degrades when the door occupies 
an intermediate angle, as the surface normal estimation becomes ambiguous and the 
YOLO bounding box does not clearly resolve to either state. Future work should explore 
incorporating door angle estimation as a continuous variable rather than a binary 
classifier, potentially using learned or geometric methods to track the door's pose 
throughout its sweep.
 
\subsection{System Timing and Real-Time Feasibility}
The timing characteristics summarized in Table~\ref{tab:timing} confirm that the 
system operates within real-time constraints appropriate for dynamic human 
environments. The principal latency contributor is the inference pipeline: offloading 
YOLOv11n to an external NVIDIA Jetson Orin NX introduces an end-to-end perception 
latency of approximately 200 ms, which is the latency bound for the BT's reactive 
behaviors. At a BT tick rate of 10 Hz, this latency represents two missed ticks in a 
worst-case scenario, which remains acceptable for the timescales associated with 
human walking speeds in doorway-width environments.
 
The IK solver's warm-starting strategy of initializing from the robot's current joint 
configuration at the start of the approach phase is critical to achieving 20 ms solve 
times. Cold-started solves on the HSR's 5-DOF arm geometry near door-button 
configurations require substantially more iterations. This warm-starting approach 
effectively transforms the IK solve into a local refinement problem, avoiding saddle 
points and reducing jitter in the resulting trajectories.
 
\subsection{Behavior Tree Composability}
The full autonomy chain scenario (S5) validates the compositional properties of the 
BT architecture across an extended operational sequence: two independent 
door-traversal instances, a manipulation subtree, and a return navigation loop, all 
governed by a single unified tree. The single failure in this scenario was attributable to 
localization drift rather than BT logic, suggesting that the executive layer is sound 
but dependent on the quality of the underlying localization stack over extended 
trajectories. This motivates future integration of loop-closure or landmark-based 
correction into the navigation pipeline for multi-room deployments.

\subsection{Limitations and Future Work}
Several limitations merit consideration. First, the YOLO models were trained on a 
controlled dataset of 600 images, which may not fully represent the variance in 
lighting, door finish, and button styles encountered in arbitrary real-world deployments. 
Data augmentation and domain randomization during training could improve 
generalization. Second, the current social preemption heuristic relies solely on 
spatial proximity rather than intent estimation; incorporating velocity and heading 
prediction for nearby humans would enable more nuanced yielding behaviors, such as 
distinguishing a human walking toward the door from one walking parallel to it. Finally, 
the binary door state model should be extended to handle the continuous door angle 
domain, enabling the robot to make probabilistic decisions under partially open conditions 
rather than defaulting to a failure mode.

\section{Conclusion}

This paper presented a modular Behavior Tree architecture for socially-aware autonomous door traversal on the Toyota Human Support Robot, motivated by the demands of assisting an emergency evacuation. The system targets two roles in that setting: keeping doors open and moving so people can follow a clear path to an exit, and retrieving payloads such as water bottles and flashlights to hand off to the responders carrying out the rescue. Across 105 trials spanning four hardware and three simulation scenarios, the system completed 97 successfully, with failures concentrated in perception rather than in planning or manipulation. These results support the framework as a practical basis for robotic assistance in evacuation settings, and motivate the perception improvements outlined above as the next step toward deployment beyond controlled environments.

\bibliographystyle{IEEEtran}
\bibliography{IEEEabrv,references}

\end{document}